\UseRawInputEncoding
\pdfoutput=1
\documentclass[11pt]{article}
\usepackage{fix-cm}
\usepackage[preprint]{acl}

\usepackage{times}
\usepackage{microtype}
\usepackage{graphicx}   % For including images
\usepackage{caption}    % For customizing captions
\usepackage{multirow}
\usepackage{tabularx}
\usepackage{booktabs}
\definecolor{citecolor}{HTML}{0071BC}
\definecolor{linkcolor}{HTML}{ED1C24}
\definecolor{LGray}{gray}{0.97}
\usepackage{multicol}
\usepackage{colortbl}
\usepackage{xcolor}
\definecolor{tabhighlight}{HTML}{e5e5e5}
\usepackage{color}
\usepackage{xspace}
\usepackage{pifont}
\usepackage{colortbl} 

\usepackage{tcolorbox}
\tcbuselibrary{listings}
\usepackage{listings}

\usepackage{graphicx}
\usepackage{tikz}
\usetikzlibrary{trees}

\definecolor{darkgreen}{rgb}{0.0, 0.5, 0.0}  % Adjust the RGB values to your preference
\newcommand{\cmark}{\textcolor{darkgreen}{\scalebox{1}[1.0]{\ding{51}}}}
\newcommand{\xmark}{\textcolor{red}{\ding{55}}}  % Cross

\lstdefinelanguage{json}{
    basicstyle=\ttfamily\scriptsize,
    numbers=left,
    numberstyle=\tiny,
    stepnumber=1,
    breaklines=true,
    showstringspaces=false,
    frame=single,
    backgroundcolor=\color{gray!10},
    keywordstyle=\color{blue},
    stringstyle=\color{red}
}

\usepackage[T1]{fontenc}
\usepackage[utf8]{inputenc}
\usepackage{inconsolata}
\usepackage{graphicx}

\title{
    \begin{minipage}{0.12\textwidth} % Adjust width for logo
        \raggedleft
        \includegraphics[height=3cm]{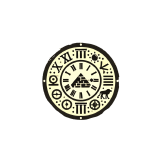} % Adjust height for logo spanning two lines
    \end{minipage}%
    \hspace{0.005cm} % Space between logo and text
    \begin{minipage}{0.8\textwidth} % Title text alignment
        \centering
        \textbf{Time Travel: A Comprehensive Benchmark to Evaluate LMMs on Historical and Cultural Artifacts}
    \end{minipage}
}

\author{\\ {Sara Ghaboura}\textsuperscript{1$\dagger$} \quad{Ketan More}\textsuperscript{1$\dagger$} \quad{Ritesh Thawkar}\textsuperscript{1}\quad {Wafa Alghallabi }\textsuperscript{1} \quad{Omkar Thawakar}\textsuperscript{1} \\
  {Fahad Shahbaz Khan}\textsuperscript{1,2}  \quad    {Hisham Cholakkal}\textsuperscript{1} \quad  
     {Salman Khan}\textsuperscript{1,3} \quad  {Rao Muhammad Anwer}\textsuperscript{1,4}\\
     \fontsize{11pt}{12pt}\selectfont \textsuperscript{1}Mohamed bin Zayed University of AI, \textsuperscript{2}Linköping University, \textsuperscript{3}Australian National University,
     \textsuperscript{4}Aalto University \\
     \fontsize{10pt}{12pt}\selectfont \{{sara.ghaboura, ketan.more, omkar.thawakar}\}@mbzuai.ac.ae \\
 {\hypersetup{urlcolor=blue}
\fontsize{11pt}{12pt}\selectfont \href{https://mbzuai-oryx.github.io/TimeTravel/}{https://mbzuai-oryx.github.io/TimeTravel/}}}

\begin{document}
\maketitle
\begin{abstract}
Understanding historical and cultural artifacts demands human expertise and advanced computational techniques, yet the process remains complex and time-intensive. While large multimodal models offer promising support, their evaluation and improvement require a standardized benchmark. To address this, we introduce \emph{TimeTravel}, a benchmark of 10,250 expert-verified samples spanning 266 distinct cultures across 10 major historical regions. Designed for AI-driven analysis of manuscripts, artworks, inscriptions, and archaeological discoveries, TimeTravel provides a structured dataset and robust evaluation framework to assess AI models’ capabilities in classification, interpretation, and historical comprehension. By integrating AI with historical research, TimeTravel fosters AI-powered tools for historians, archaeologists, researchers, and cultural tourists to extract valuable insights while ensuring technology contributes meaningfully to historical discovery and cultural heritage preservation. We evaluate contemporary AI models on TimeTravel, highlighting their strengths and identifying areas for improvement. Our goal is to establish AI as a reliable partner in preserving cultural heritage, ensuring that technological advancements contribute meaningfully to historical discovery. Our code is available at: \href{https://github.com/mbzuai-oryx/TimeTravel}{https://github.com/mbzuai-oryx/TimeTravel}.
\def\thefootnote{$\dagger$}\footnotetext{Equal contribution.}
\end{abstract}

\section{Introduction}
In recent years, Large Multimodal Models (LMMs) have made significant strides in visual reasoning, perception, and multimodal understanding. Models such as GPT-4V \cite{openai2024gpt4ocard} and LLaVA \cite{liu2023llava} have excelled in image captioning, visual question answering (VQA), and complex visual reasoning, driving the development of benchmarks~\cite{chiu2024culturalbench,nayak2024benchmarking,alwajih2024peacock} to assess their capabilities. These benchmarks predominantly focus on modern objects, cultural landmarks, and textual sources, extending multimodal AI applications to domains such as medical imaging, remote sensing, and real-world scene understanding~\cite{ghaboura2024camel}. However, a critical gap remains—LMMs fail to address the historical dimension of visual data, particularly artifacts that shaped human civilization.

\begin{figure}[t!]
    \centering
    \includegraphics[width=\columnwidth]{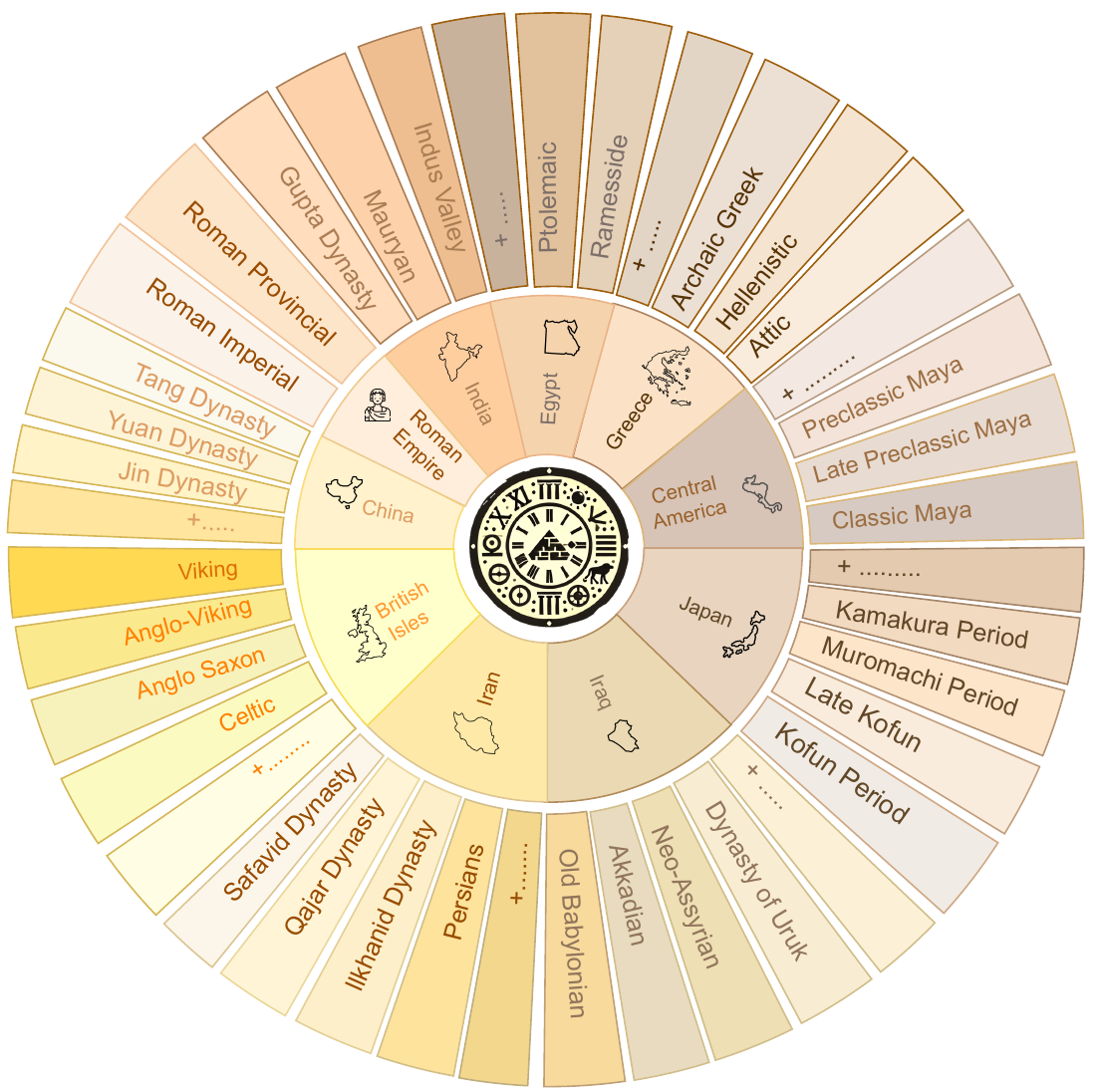}
    \caption{TimeTravel Taxonomy categorizes artifacts from 10 major civilizations, spanning diverse historical and prehistoric periods. It encompasses 266 distinct cultures and over 10k manually verified historical artifact samples, providing a structured framework for comprehensive AI-driven analysis.}
    \label{fig:taxonomy}
    \vspace{-1.5em}
\end{figure}

\begin{figure*}[t!]
\centering  
\includegraphics[width=\textwidth,height=4cm]{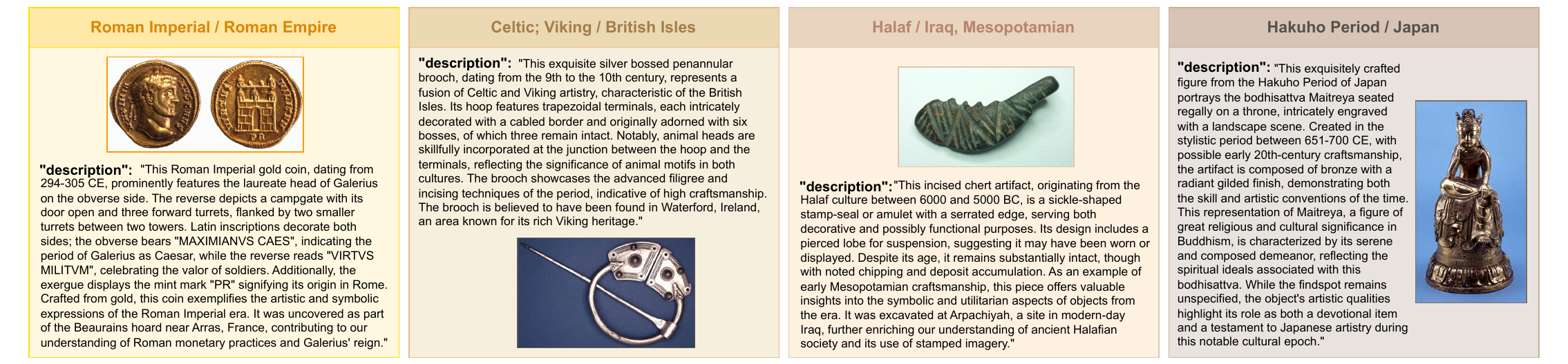}
\vspace{-2em}
  \caption{
  % \small
    \textbf{TimeTravel Samples.} Showcasing diverse cultural representations from various regions across the globe, these examples span multiple artifact categories, including coins, accessories, tools, and statues from ancient civilizations. Each artifact is accompanied by a detailed description, providing valuable contextual and historical insights. Additional TimeTravel examples can be found in Fig.\ref{fig:appendix_qual_examples} and Fig.\ref{fig:appendix_data_examples}.
  }
  \label{fig:samples_fig}
  \vspace{-1em}
\end{figure*}

Historical artifacts, from ancient manuscripts and inscriptions to architectural ruins and cultural symbols, offer invaluable insights into the evolution of societies, artistic expression, and technological advancements. These artifacts preserve cultural heritage and serve as primary sources for understanding belief systems, trade networks, and socio-political structures of past civilizations. However, interpreting them requires deep contextual knowledge, which current LMMs struggle to achieve, particularly in non-English and non-Western historical contexts. While some models have been extended to low-resource languages to bridge cultural gaps~\cite{heakl2025ainarabicinclusivelarge}, they lack systematic capabilities to analyze artifacts from diverse civilizations. This limitation highlights the urgent need for a specialized benchmark that evaluates AI’s ability to process and understand historical artifacts with cultural and temporal awareness.\\
To address this challenge, we introduce TimeTravel, an open-source comprehensive benchmark (see Table~\ref{tab:dataset_comparison}) for evaluating LMM performance in historical artifact analysis across diverse civilizations. TimeTravel encompasses several major ancient and prehistoric civilizations across 10 distinct regions, spanning 266 cultural groups. It offers a structured taxonomy tailored for AI-driven historical research (see Fig.~\ref{fig:taxonomy}). Unlike existing benchmarks that focus on generic object recognition, TimeTravel prioritizes historical knowledge, contextual reasoning, and cultural preservation, making it a pioneering effort in multimodal AI evaluation. The benchmark consists of over 10k curated samples, each accompanied by high-quality images of manuscripts, inscriptions, sculptures, paintings, and archaeological discoveries. These samples assess key aspects of multimodal understanding, including visual perception, contextual reasoning, and cross-civilizational knowledge. Meticulously verified by historians and archaeologists, the dataset ensures accuracy, cultural relevance, and historical integrity. By evaluating both closed- and open-source LMMs on TimeTravel, we aim to identify their strengths and limitations in handling historically significant artifacts, paving the way for AI models that contribute meaningfully to cultural heritage preservation and historical analysis.

\begin{table}[ht]
\centering
\resizebox{\columnwidth}{!}{
\begin{tabular}{l|ccccccc}
\toprule
\multirow{2}{*}{\textbf{Domain}} & \textbf{British} & \multirow{2}{*}{\textbf{MMMU}} & \textbf{Oracle-} & \multirow{2}{*}{\textbf{Ithaca}} & \textbf{Kao} & \textbf{HUST-} & \textbf{TimeTravel} \\
 &\textbf{Museum} & &\textbf{MNIST} & & \textbf{Kore}&\textbf{OBS} & \textbf{(ours)}\\
\midrule
Hist. Artifact Recog. & \cmark & \xmark & \xmark & \xmark & \cmark & \xmark & \cmark \\
Geographic Region & \cmark & \xmark & \xmark & \cmark & \cmark & \xmark & \cmark  \\
Ancient Artifacts & \cmark & \xmark & \xmark & \xmark & \xmark & \xmark & \cmark  \\
Contextual History & \xmark & \xmark & \xmark & \xmark & \xmark & \xmark & \cmark  \\
Image-Text Pairs & \cmark & \cmark & \xmark & \xmark & \cmark & \cmark & \cmark  \\
Open-Source & \xmark & \cmark & \cmark & \xmark & \cmark & \cmark & \cmark  \\
\bottomrule
\end{tabular}
\vspace{-1em}
}
\caption{
% \small
The comparison of datasets and benchmarks for historical and cultural artifacts, evaluating features like \textbf{artifact recognition}, \textbf{geographic coverage}, \textbf{multimodal understanding}, and \textbf{metadata inclusion} with existing data such as British Museum~\cite{tully2020british}, MMMU~\cite{yue2023mmmu}, Oracle-MNIST~\cite{wang2022oracle}, Ithaca~\cite{Assael2022RestoringAA}, KaoKore~\cite{tian2020kaokore}, HUST-OBS~\cite{wang2024open}. 
TimeTravel stands out as the most comprehensive benchmark, uniquely integrating multimodal data, historical context, and a dedicated focus on ancient artifacts to support AI-driven cultural heritage research.
% TimeTravel excels as the most comprehensive benchmark, uniquely combining multimodal data, historical context, and a focus on ancient artifacts to support AI-driven cultural heritage research.
}
\label{tab:dataset_comparison}
\vspace{-1em}
\end{table}

\begin{figure*}[t!]
\centering
  \includegraphics[width=\textwidth,height=4.5cm]{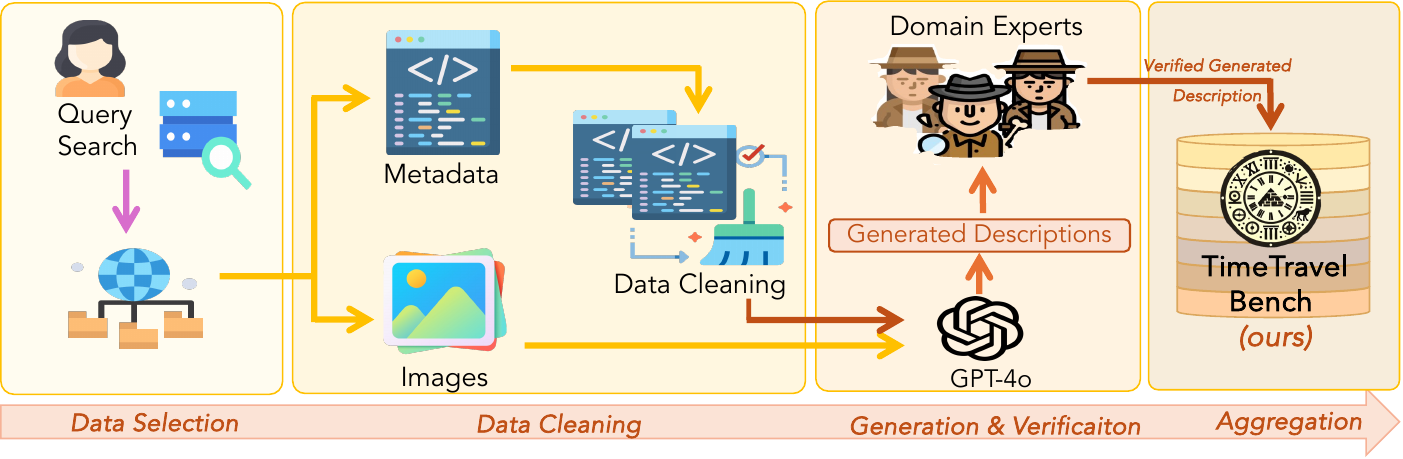}
  % \vspace{-0.3em}
  \caption{
  % \small
 \textbf{TimeTravel Data Pipeline.} A structured workflow that collects image and text data from museum websites, cleans metadata, and integrates it with visual content. The GPT-4o model generates detailed, context-aware descriptions, which are refined by experts for accuracy before forming the TimeTravel Benchmark.
  }
  \label{fig:data_pipeline}
  \vspace{-1em}
\end{figure*}

\section{The TimeTravel Dataset}

\subsection{Data Collection}
Our research is based on a well-structured and meticulously curated dataset sourced from museum collections, which houses an extensive collection of artifacts from diverse civilizations. From this vast repository, we compiled a dataset spanning 266 cultural groups, allowing the analysis of cultural, technological, and social developments over a broad historical timeline.

To ensure the integrity of our benchmark, we followed a systematic data collection process. We first identified key civilizations and historical periods relevant to our study, then collaborated closely with experts to validate the authenticity and completeness of each record. As a result, our dataset comprises 10,250 carefully curated samples (see Fig~\ref{fig:samples_fig}). Each entry—ranging from artifacts and inscriptions to ancient manuscripts—was meticulously verified by historians and archaeologists, ensuring accuracy and reliability. By incorporating data from multiple civilizations, our benchmark provides a diverse and comprehensive perspective, avoiding the limitations of a single historical narrative while preserving the historical context for in-depth analysis. This meticulous approach allows us to reveal significant patterns in human history, offering valuable insights into the evolution of human history and civilizations over time. 

\subsection{Image-Text pair Generation}
The dataset features a diverse range of historical objects, ensuring comprehensive documentation and contextual understanding. However, many metadata fields—such as title, iconography, and date—were missing or incomplete. To address this, we employed GPT-4o to generate detailed, context-aware textual descriptions based on the available metadata (see Fig.~\ref{fig:raw-data-india} and ~\ref{fig:raw-data-maya}). To further enhance usability, we structured these descriptions into image-text pairs, ensuring that each artifact is not only visually documented but also enriched with contextual and cultural insights. By improving multimodal model compatibility and supporting digital archiving, this approach strengthens research in cultural heritage preservation while bridging gaps in existing records.

\subsection{Data Filtering and Verification}
To guarantee the accuracy and reliability of our dataset, we implemented a rigorous data filtering and verification process (Fig.~\ref{fig:data_pipeline}). This process combined manual expert validation with automated techniques to eliminate inconsistencies, fill in missing details where possible, and authenticate historical records. During data cleaning, we addressed missing or incomplete metadata—such as titles, dates, and iconography—by cross-referencing museum archives, academic sources, and expert insights. Unavailable key information was transparently documented. Additionally, automated checks identified formatting inconsistencies, metadata mapping errors, and numerical anomalies, ensuring a structured and standardized dataset.
For verification, we collaborated with historians, archaeologists, and museum curators to review each artifact’s description, cultural attribution, and historical significance. Expert validation ensured that generated textual descriptions were accurate, contextually relevant, and aligned with historical records. This rigorous process enhances the dataset’s credibility, making it a valuable resource for historical research, machine learning, and cultural heritage preservation while ensuring reliable insights into human history. Additional details are presented in Appendix (Sec.~\ref{appendix:data_stats}).

\section{TimeTravel Benchmark Evaluation}
\begin{table*}[t!]
\centering
\small
\setlength{\tabcolsep}{7pt}
    \resizebox{\textwidth}{!}{%
    \begin{tabular}{llcccccc}
    \toprule
    % \hline
    \rowcolor{brown!25} & \textbf{Model} & \textbf{BLEU} & \textbf{METEOR} & \textbf{ROUGE-L} & \textbf{SPICE} & \textbf{BERTScore} & \textbf{LLM-Judge} \\
    \midrule
    \multirow{4.2}{*}{
    \begin{tabular}[c]{@{}c@{}}
    \rotatebox{90}{\small{Closed}}\end{tabular}}
    % \rowcolor{brown!15}\multicolumn{7}{l}{\textbf{Close-Source}} \\
    & GPT-4o-0806~\cite{openai2024gpt4ocard} & \textbf{0.1758} & 0.2439 & \textbf{0.1230} & \textbf{0.1035} & \textbf{0.8349} & \textbf{0.3013} \\
    & Gemini-2.0-Flash~\cite{gemini1.5} & 0.1072 & 0.2456 & 0.0884 & 0.0919 & 0.8127 & 0.2630 \\
    & Gemini-1.5-Pro~\cite{gemini1.5} & 0.1067 & 0.2406 & 0.0848 & 0.0901 & 0.8172 & 0.2276 \\
    & GPT-4o-mini-0718~\cite{gpt4omini} & 0.1369 & \textbf{0.2658} & 0.1027 & 0.1001 & 0.8283 & 0.2492 \\
    \midrule
    \multirow{3.2}{*}{
    \begin{tabular}[c]{@{}c@{}}
    \rotatebox{90}{\small{Open}}\end{tabular}}
    % \rowcolor{brown!15}\multicolumn{7}{l}{\textbf{Open-Source}} \\
    & Llama-3.2-Vision-Inst~\cite{llama3.2} & 0.1161 & 0.2072 & 0.1027 & 0.0648 & 0.8111 & 0.1255 \\
    & Qwen-2.5-VL~\cite{Qwen2.5-VL} & 0.1155 & 0.2648 & 0.0887  & 0.1002 & 0.8198 & 0.1792 \\
    & Llava-Next~\cite{liu2024llavanext} & 0.1118 & 0.2340 & 0.0961 & 0.0799 & 0.8246 & 0.1161 \\
    \bottomrule
    \end{tabular}
    }
    \vspace{-1em}
\caption{
% \small
Performance comparison of various closed and open-source models on our proposed TimeTravel benchmark. 
}
\label{tab:model_performance}
\end{table*}

\begin{table*}[t!]
\centering
\small
\setlength{\tabcolsep}{7pt}
    \resizebox{\textwidth}{!}{%
    \begin{tabular}{llcccccccccc}
    \toprule
     \rowcolor{brown!25} & \textbf{Model}  & \textbf{India} & \textbf{Roman} & \textbf{China} & \textbf{British} & \textbf{Iran} & \textbf{Iraq} & \textbf{Japan} & \textbf{Central} & \textbf{Greece} & \textbf{Egypt} \\
    \rowcolor{brown!25} & & & \textbf{Empire} & & \textbf{Isles} & & & & \textbf{America} & & \\
    \midrule
    \multirow{4.2}{*}{
    \begin{tabular}[c]{@{}c@{}}
    \rotatebox{90}{\small{Closed}}\end{tabular}}
    & GPT-4o-0806& \textbf{ 0.2491} & \textbf{0.4463} & \textbf{0.2491} & \textbf{0.1899} & \textbf{0.3522} & \textbf{0.3545} & \textbf{0.2228} &\textbf{ 0.3144} &\textbf{ 0.2757 }& \textbf{0.3649} \\
    & Gemini-2.0-Flash & 0.1859 & 0.3358 & 0.2059 & 0.1556 & 0.3376 & 0.3071 & 0.2000 & 0.2677 & 0.2582 & 0.3602 \\
    & Gemini-1.5-Pro & 0.1118 & 0.2632 & 0.2139 & 0.1545 & 0.332 & 0.2587 & 0.1871 & 0.2708 & 0.2088 & 0.2908 \\
    & GPT-4o-mini-0718 & 0.2311 & 0.3612 & 0.2207 & 0.1866 & 0.2991 & 0.2632 & 0.2087 & 0.3195 & 0.2101 & 0.2501 \\
    \midrule
    \multirow{3.2}{*}{
    \begin{tabular}[c]{@{}c@{}}
    \rotatebox{90}{\small{Open}}\end{tabular}}
    & Llama-3.2-Vision-Inst & 0.0744 & 0.1450 & 0.1227 & 0.0777 & 0.2000 & 0.1155 & 0.1075 & 0.1553 & 0.1351 & 0.1201 \\
    & Qwen-2.5-VL & 0.0888 & 0.1578 & 0.1192 & 0.1713  & 0.2515 & 0.1576 & 0.1771 & 0.1442 & 0.1442 & 0.2660 \\
    & Llava-Next & 0.0788 & 0.0961 & 0.1455 & 0.1091 & 0.1464 & 0.1194 & 0.1353 & 0.1917 & 0.1111 & 0.0709 \\
    \bottomrule
    \end{tabular}
    }
    \vspace{-1em}
\caption{
% \small
Analysis of LLM-Judge evaluation of various models in describing archaeological artifacts across civilizations from different geographical locations. Additional comparisons are presented in Appendix (Table~\ref{tab:model_performance_regions2}).
}
\vspace{-1em}
\label{tab:model_performance_region1}
\end{table*}

\noindent\textbf{Evaluation Metric:}
To assess the quality, accuracy, and relevance of our generated textual descriptions, we employed a combination of traditional and advanced metrics. BLEU~\cite{papineni2002bleu} and ROUGE-L~\cite{lin2004rouge} evaluate linguistic fluency and structural similarity, ensuring syntactic alignment with reference texts. METEOR~\cite{banerjee2005meteor} enhances this by incorporating synonym matching and paraphrasing, improving adaptability to human variations. 
% CIDEr~\cite{vedantam2015cider} captures relevance and consensus by measuring alignment with human expectations, while 
SPICE~\cite{anderson2016spice} assesses semantic accuracy through scene graph analysis, preserving object relationships and cultural context. Additionally, BERTScore~\cite{zhang2019bertscore} offers a deep learning-based evaluation of semantic similarity, capturing contextual meaning beyond simple word overlap. LLM-Judge further enhances assessment by evaluating coherence, factual accuracy, and contextual appropriateness.

\noindent\textbf{Results and Analysis:}
Our evaluation of closed-source and open-source models on the TimeTravel dataset reveals clear differences in their ability to generate historically accurate descriptions (see Table~\ref{tab:model_performance}). Among closed-source models, GPT-4o-0806 achieved the highest BLEU (0.1758), ROUGE-L (0.1230), SPICE (0.1035), BERTScore (0.8349), and LLM-Judge score (0.3013), indicating superior semantic alignment and contextual richness. However, its lower METEOR score (0.2439) suggests that while it generates highly structured descriptions, they may lack word-level diversity and fluency. GPT-4o-mini-0718, despite scoring slightly lower in BLEU (0.1369) and ROUGE-L (0.1027), outperformed all models in METEOR (0.2658), highlighting its strength in producing more lexically diverse and well-formed outputs. Gemini-2.0-Flash and Gemini-1.5-Pro, while achieving moderate performance across all metrics, demonstrated weaker lexical alignment (BLEU: 0.1072, 0.1067) and semantic coherence (BERTScore: 0.8127, 0.8172), suggesting that they may struggle with historical specificity and structured descriptions.
% Among open-source models, Qwen-2.5-VL performed best, balancing fluency and semantic coherence. However, Llama-3.2-Vision-Inst and Llava-Next struggled with lower SPICE and ROUGE-L scores, indicating difficulties in capturing object relationships and historical context. While closed-source models currently lead in accuracy and contextual richness, open-source alternatives are steadily improving.
Among open-source models, Qwen-2.5-VL performed the best, achieving higher BLEU (0.1155), METEOR (0.2648), and SPICE (0.1002) compared to its counterparts. These scores indicate a better balance between fluency and contextual accuracy, making it a strong contender despite being an open-source model. Llama-3.2-Vision-Inst and Llava-Next, however, showed lower SPICE (0.0648, 0.0799) and LLM-Judge scores (0.1255, 0.1161), suggesting difficulties in capturing object details and historical context. 
% While closed-source models currently outperform open-source alternatives in overall accuracy and contextual relevance, the steady improvement of open-source models indicates promising potential for future fine-tuning and dataset expansion to bridge the performance gap.

Table~\ref{tab:model_performance_region1} presents the LLM-Judge evaluation of models in describing archaeological artifacts across civilizations from different geographic regions. GPT-4o-0806 outperformed other models in describing archaeological artifacts, excelling in regions like the Roman Empire, Iran, Iraq, and Egypt, indicating strong contextual understanding. GPT-4o-mini-0718 and Gemini-2.0-Flash showed strengths in India, Central America, and China, but with some limitations. Among open-source models, Qwen-2.5-VL performed best in Iran, the British Isles, and Egypt, though overall, closed-source models provided more accurate historical descriptions. Additional analysis based on the METEOR score is presented in Appendix (Table~\ref{tab:model_performance_regions2}).
% Open models show potential but require further fine-tuning and richer training data to improve their understanding of global cultural heritage.

Overall, closed-source models outperform open-source models in generating context-aware descriptions, but ongoing improvements in open-source models highlight opportunities for fine-tuning and dataset expansion. These findings will guide further model enhancements, advancing AI-driven historical analysis and cultural heritage preservation.

\section{Conclusion}
We present the TimeTravel dataset, a curated collection of historical artifacts from 10 cultural regions, extensively curated by domain experts. We developed a rigorous data collection, filtering, and verification process, ensuring accuracy and completeness. Using GPT-4o, we generated detailed textual descriptions, making the dataset more accessible and valuable for AI-driven historical research. Our evaluation, using BLEU, METEOR, ROUGE-L, CIDEr, SPICE, BERTScore, and LLM-Judge, showed that closed-source models outperformed open-source alternatives, though open models are rapidly improving. Our analysis highlights the potential of LMMs in bridging gaps in historical records while maintaining academic integrity. By leveraging AI-driven methodologies, this work sets the foundation for advancing cultural heritage preservation and enhancing digital humanities research, ensuring greater accessibility and accuracy in historical documentation.

\section{Limitations and Societal Impact}
While this research demonstrates the potential of LMMs in enhancing historical documentation, the quality of generated descriptions depends on the completeness and accuracy of the input data. In cases where historical records are fragmented or ambiguous, AI-generated text may lack full contextual depth. Additionally, biases present in training data can influence how models interpret and describe cultural artifacts, necessitating continuous evaluation and expert validation to ensure historical accuracy and cultural sensitivity. Despite these challenges, this research contributes to cultural heritage preservation, educational accessibility, and AI-driven humanities research. By digitizing and enriching historical records, it enables wider public engagement with history, supports museum digitization efforts, and provides a foundation for future advancements in AI-assisted historical analysis, bridging the gap between technology and human expertise in understanding our collective past.

\bibliography{arxiv}

\clearpage
\appendix

\section{Appendix}
\label{sec:appendix}

In this appendix, we provide additional details to support our research, including related work, data statistics, and a comprehensive overview of archaeological samples from various cultures, civilizations, and dynasties. The related work section provides a review of existing research in AI-driven historical text generation, contextualizing our contributions within the broader field.  The data statistics section offers a structured breakdown of collected samples, highlighting their geographical distribution and cultural significance. Additionally, the inclusion of archaeological records from diverse historical periods reinforces the depth and diversity of the dataset. 
% This supplementary information enhances the transparency and reproducibility of our research while providing a broader perspective on historical narratives across different regions and time periods.

\begin{table*}[t!]
\centering
\small
\setlength{\tabcolsep}{7pt}
    \resizebox{\textwidth}{!}{%
    \begin{tabular}{lcccccccccc}
    \toprule
    \rowcolor{brown!25}\textbf{Model} & \textbf{India} & \textbf{Roman} & \textbf{China} & \textbf{British} & \textbf{Iran} & \textbf{Iraq} & \textbf{Japan} & \textbf{Central} & \textbf{Greece} & \textbf{Egypt} \\
    \rowcolor{brown!25} & & \textbf{Empire} & & \textbf{Isles} & & & & \textbf{America} & & \\
    \midrule
    % \hline
    % \rowcolor{brown!15}\multicolumn{11}{l}{\textbf{Close-Source}} \\
    GPT-4o-0806~\cite{openai2024gpt4ocard} & 0.2566 & 0.2713 & 0.2324 & 0.2175 & 0.2486 & 0.2428 & 0.2269 & 0.2384 & 0.2441 & 0.2567 \\
    Gemini-2.0-Flash~\cite{gemini1.5} & 0.2478 & 0.2603 & 0.2183 & 0.2189 & 0.2432 & 0.242 & 0.2256 & 0.2264 & 0.2488 & 0.2588 \\
    Gemini-1.5-Pro~\cite{gemini1.5} & 0.2586 & 0.2596 & 0.2198 & 0.2203 & 0.2535 & 0.2524 & 0.2253 & 0.2218 & 0.2551 & 0.268 \\
    GPT-4o-mini-0718~\cite{gpt4omini} &\textbf{ 0.2762} & 0.2731 &\textbf{ 0.2570} & \textbf{ 0.2531} &\textbf{ 0.2660} & \textbf{0.2640 }& \textbf{0.2611} & \textbf{0.2741} & 0.2649 & 0.2741 \\
    \midrule
    % \hline
    % \rowcolor{brown!15}\multicolumn{11}{l}{\textbf{Open-Source}} \\
    Llama-3.2-Vision-Inst~\cite{llama3.2} & 0.2128 & 0.2253 & 0.1867 & 0.1917 & 0.2115, & 0.2078 & 0.1944 & 0.1979 & 0.2138 & 0.2182 \\
    Qwen-2.5-VL~\cite{Qwen2.5-VL} & 0.2707 &\textbf{ 0.2815} & 0.2526 & 0.2464  & 0.2607 & 0.2631 & 0.2499 & 0.2587 & \textbf{0.2713} & \textbf{0.2827} \\
    Llava-Next~\cite{liu2024llavanext} & 0.2482 & 0.2527 & 0.2156 & 0.2192 & 0.2389 & 0.2321 & 0.2207 & 0.2196 & 0.2388 & 0.2427 \\
    \bottomrule
    \end{tabular}
    }
\caption{
% \small
Analysis of METEOR Evaluation of various models in describing archaeological artifacts across civilizations from different geographical regions. 
}
\label{tab:model_performance_regions2}
\end{table*}

\section{Related Work}
Recent years have seen significant progress in studying cultural representation in AI, particularly in behavioral patterns, food, landmarks, and historical knowledge. However, most works focus on misalignment and biases in AI models or modern cultural trends, rather than positioning artifacts within their historical context and era across ancient civilizations. Meanwhile, studies on cultural inclusion in LLMs highlight the challenges of capturing the contextual and multifaceted nature of culture, emphasizing the limitations of text-based models in representing underrepresented cultures and the need for more robust evaluation methods \cite{adilazuarda2024towards}.

% \subsection{Cultural Bias in Language and Vision Models}
Research on cultural influences in AI has increasingly focused on biases and misalignment in language models, particularly how they reflect and perpetuate dominant cultural norms. Early research on cultural biases in LLMs revealed their alignment with Western norms, particularly in moral reasoning, historical narratives, and societal values. Ramezani et al. (2023) analyze how monolingual English language models tend to reflect Western moral norms more strongly than diverse cultural perspectives, limiting their applicability in cross-cultural ethical contexts \cite{Ramezani2023KnowledgeOC}. Tao et al. (2024) further highlight the overrepresentation of Anglo-American and Protestant European values in AI-generated content, often underrepresenting non-Western traditions and belief systems \cite{tao2024cultural}. Similarly, Bu et al. (2025) explore value misalignment in cultural heritage-related text generation, warning of historical inaccuracies, cultural identity erosion, and oversimplification of complex narratives, with 65\% of the generated content showing significant misalignment \cite{bu2025investigation}.

To mitigate these biases, several approaches have been proposed. AlKhamissi et al. (2024) introduce Anthropological Prompting, a method that encourages LLMs to reason like cultural anthropologists by incorporating both emic (insider) and etic (outsider) perspectives \cite{AlKhamissi2024InvestigatingCA}. Similarly, Li et al. (2024) propose CultureLLM, a fine-tuning approach designed to integrate cultural knowledge into LLMs, particularly for low-resource cultures \cite{li2024culturellm}. While these techniques improve cultural alignment, their focus remains on modern cultural settings, leaving gaps in historical artifact contextualization across different time periods.

With the rise of Vision-Language Models (VLMs), cultural research has expanded to multimodal AI, revealing similar biases. Liu et al. (2025) introduce CultureVLM, a model designed to improve cultural understanding in VLMs, highlighting their inability to recognize non-Western cultural symbols, historical artifacts, and traditional gestures \cite{liu2025culturevlm}. Their work also presents CultureVerse, a large-scale multimodal dataset covering several cultural concepts, designed to evaluate VLMs' cultural reasoning. However, CultureVerse has a primary focus on modern cultural symbols, traditions, and everyday life. Additionally, Romero et al. (2024) develop CVQA, a multilingual and culturally diverse Visual Question Answering (VQA) benchmark, which reveals that state-of-the-art VLMs struggle with culturally grounded reasoning, particularly in non-Western contexts \cite{romero2024cvqa}. However, these datasets primarily focus on present-day cultural contexts, even when historical artifacts are included, as they are often framed through the lens of modern nations rather than their original civilizations and historical epochs \cite{liu2025culturevlm}. This leaves a significant gap in representing artifacts within their authentic temporal and cultural contexts.

Efforts to bridge AI research with historical studies have led to the development of Historical Large Language Models (HLLMs), trained on historical texts to simulate past societies' psychology and value systems \cite{varnum2024large}. These models aim to provide insight into long-term cultural evolution, but their reliance on text-only representations limits their application in multimodal historical studies. Similarly, Assael et al. (2022) introduce Ithaca, a deep learning model designed to assist historians in restoring, geographically attributing, and dating ancient Greek inscriptions, significantly improving accuracy over traditional methods \cite{Assael2022RestoringAA}. While these works contribute to historical AI, they primarily focus on text-based reconstruction rather than multimodal representations of historical artifacts across civilizations.

TimeTravel fills this gap by providing a 10k historical artifact open-source dataset spanning 10 ancient world regions (prehistoric and historic), offering the first benchmark to evaluate LMMs on temporal-cultural understanding with expert verification. Unlike prior datasets focused on contemporary cultural knowledge, TimeTravel enables AI models to contextualize artifacts within their historical era, ensuring a more accurate representation of civilizations and their material culture. With domain expert verification, the dataset enhances reliability and authenticity, mitigating potential biases and inaccuracies in AI-generated interpretations. By integrating both textual and multimodal perspectives, TimeTravel advances research in historical-cultural AI, enabling AI systems to better understand and reason about artifacts in their original context.

\section{TimeTravel Samples Regional Distribution}
\begin{figure}[ht]
\centering
  \includegraphics[width=0.9\columnwidth]{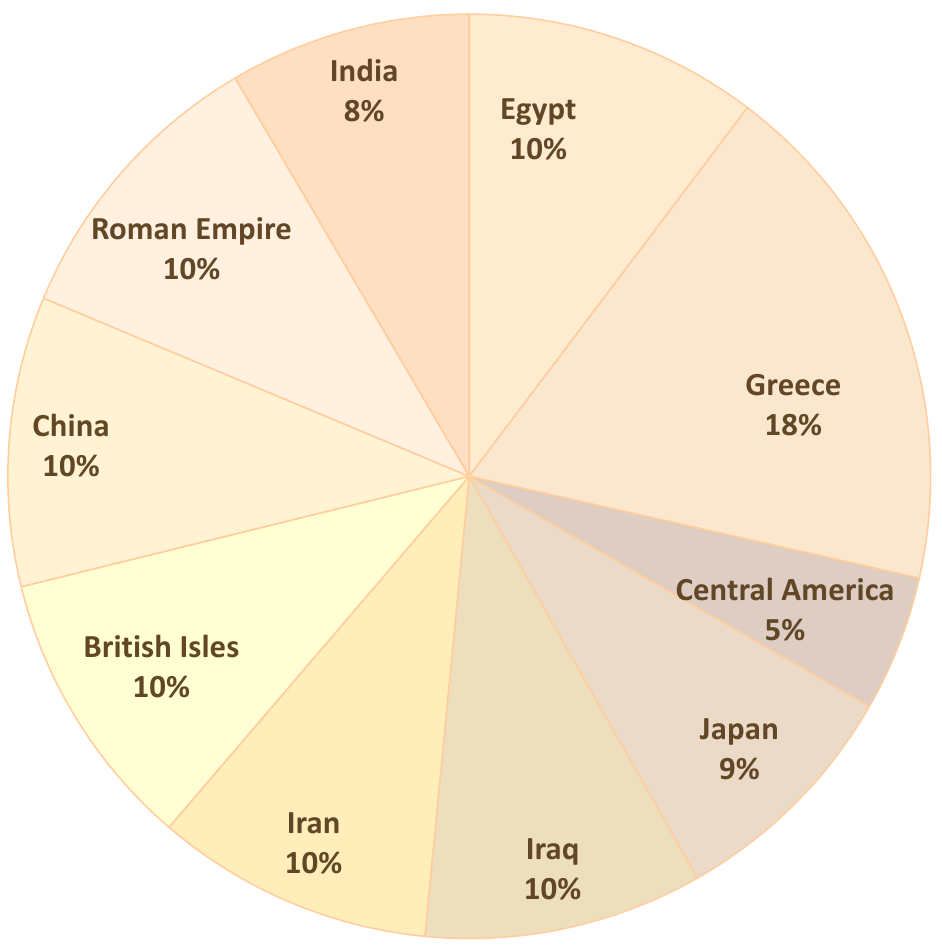}
  \caption{Regional distribution of dataset samples based on their archaeological provenance. Greece holds the largest share at 18\%,  with a balance-like distribution over regions.}
  \label{fig:sample_pie}
\end{figure}

Fig.~\ref{fig:sample_pie} illustrates the balanced regional distribution of dataset samples based on archaeological provenance. Greece holds the largest share at 18\%, followed by multiple regions, including the Roman Empire, China, British Isles, Egypt, Iraq, and Iran, each at 10\%. Japan (9\%), India (8\%), and Central America (5\%) contribute smaller yet significant portions. Overall, the dataset ensures diverse cultural representation without dominance by any single region.

Tables ~\ref{tab:culture-sample-rom} to ~\ref{tab:culture-sample-iraq} present further details about sample counts categorized by region of discovery, section, and cultural affiliation. \\
The covered areas in our study are ordered as follows:\\
Tab.~\ref{tab:culture-sample-rom} $\rightarrow$ ``Roman Empire'', Tab.~\ref{tab:culture-sample-greece} $\rightarrow$ ``Greece'', Tab.~\ref{tab:culture-sample-viking} $\rightarrow$ ``British Isles'', 
Tab.~\ref{tab:culture-sample-maya}$ \rightarrow$ ``Central America'', Tab.~\ref{tab:culture-sample-egypt} $\rightarrow$ ``Egypt'', Tab.~\ref{tab:tab:culture-sample-ind} $\rightarrow$ ``India'', \\
Tab.~\ref{tab:culture-sample-persian} $\rightarrow$ ``Iran'', Tab.~\ref{tab:culture-sample-tang} $\rightarrow$ ``China'', \\Tab.~\ref{tab:culture-sample-japan} $\rightarrow $``Japan'', and Tab.~\ref{tab:culture-sample-iraq} $\rightarrow$ ``Iraq'',

\begin{table}[ht!]
    \centering
    \small
    \begin{tabularx}{\columnwidth}{Xl}
        \hline
        \rowcolor{brown!30}\textbf{Place} & Roman Empire \\

        \rowcolor{brown!20}\textbf{Section} & Roman \\
        \hline
        \rowcolor{brown!10}\textbf{Culture} & \textbf{Samples} \\
        \hline
        Roman Imperial & 610 \\
        Roman & 3 \\
        Roman Provincial & 436 \\
        \hline
        \rowcolor{brown!10}\textbf{Total}&\textbf{1049}\\
         \hline
    \end{tabularx}
    \caption{Culture Sample Counts from the Roman Empire.}
    \label{tab:culture-sample-rom}
\end{table}

\begin{table*}[ht!]
    \centering
    \small
    \resizebox{\textwidth}{!}{
    \begin{tabularx}{\textwidth}{lllX}
        \hline
        \rowcolor{brown!30}\textbf{Place} & &Greece&\\
        \rowcolor{brown!20}\textbf{Section} & &Greek& \\
        \hline
        \rowcolor{brown!10}\textbf{Culture} & \textbf{Sample} & \textbf{Culture} & \textbf{Sample} \\
        \hline
        Greek; Hellenistic; Roman Imperial & 4 &  Hellenistic; Roman Imperial & 2 \\
        Attic & 806 &  Middle Corinthian & 5 \\
        Corinthian & 41 &  East Greek; Classical Greek & 1 \\
        Attic; Classical Greek & 47 &  Transitional Corinthian & 1 \\
        Middle Corinthian; Late Corinthian; Archaic Greek & 7 &  Classical Greek; Attic & 2 \\
        Proto-Corinthian & 4 &  Classical Greek; Attic; Archaic Greek & 1 \\
        Orientalising Period & 14 &  East Greek Archaic II; Archaic Greek & 1 \\
        Archaic Greek; Classical Greek & 1 & Attic; Western Greek & 1 \\
        Archaic Greek & 40 & East Greek & 23 \\
        Late Corinthian; Archaic Greek & 11 &  Attic; Archaic Greek & 318 \\
        Western Greek; Hellenistic & 1 &  Attic; Archaic Greek; Classical Greek & 12 \\
        Early Corinthian & 8 & Attic; Classical Greek; Archaic Greek & 3 \\
        Laconian; Archaic Greek & 10 &  Archaic Greek; East Greek & 2 \\
        Classical Greek; Corinthian; Hellenistic & 1 &  Rhodian & 3 \\
        Late Helladic IIIB & 2 &  Greek; Classical Greek & 2 \\
        Transitional Corinthian; Archaic Greek & 1 &  Early Corinthian; Archaic Greek & 3 \\
        East Greek; Hellenistic & 2 & Middle Corinthian; Archaic Greek & 11 \\
        Late Geometric IIA; Attic & 1 & East Greek; Orientalising Period & 1 \\
        Archaic Greek; Attic & 8 & Late Minoan I; Late Minoan II & 1 \\
        Late Minoan I & 2 & Archaic Greek; East Greek; North Ionian & 1 \\
        Paestan & 1 &  East Greek; Archaic Greek & 237 \\
        Early Corinthian; Middle Corinthian; Archaic Greek & 1 & Greek; Hellenistic & 2 \\
        Archaic Greek; East Dorian & 1  & Greek & 3 \\
        Hellenistic & 110  & Western Greek & 5 \\
        East Greek; Archaic Greek; Classical Greek & 1  & Roman; Hellenistic & 3 \\
        East Dorian; Archaic Greek & 2  & Classical Greek & 38 \\
        East Greek; East Dorian; Archaic Greek & 11  & Boeotian & 25 \\
        Geometric Greek; Early Proto-Attic & 1 & Hellenistic; Classical Greek & 2 \\
        East Greek; South Ionian & 1 & Geometric Greek & 8 \\
        Greek; Classical Greek; Hellenistic & 5  & Hellenistic; Roman & 4 \\
        \hline
        \rowcolor{brown!10}\textbf{Total} & & &\textbf{1869}\\
        \hline
    \end{tabularx}}
    \caption{Culture Sample Counts from Greece (Greek Section).}
    \label{tab:culture-sample-greece}
\end{table*}

 \begin{table}[ht]
    \centering
    \small
    \begin{tabularx}{\columnwidth}{Xl}
        \hline
        \rowcolor{brown!30} \textbf{Place} & British Isles \\
        \rowcolor{brown!20} \textbf{Section} & Viking \\
        \hline
        \rowcolor{brown!10} \textbf{Culture} & \textbf{Samples} \\
        \hline
        Viking; Carolingian; Late Anglo-Saxon & 1 \\
        Viking; Early Anglo-Saxon; Mid. Anglo-Saxon & 1 \\
         Middle Anglo-Saxon 
        Viking; Anglo-Saxon & 1 \\
        Celtic; Viking & 14 \\
        Viking; Late Anglo-Saxon & 19 \\
        Viking; Finno-Ugrian & 1 \\
        Anglo-Viking & 52 \\
        Viking & 895 \\
        Carolingian; Viking & 1 \\
        Viking; Medieval & 1 \\
        Late Anglo-Saxon; Viking & 1 \\
        Viking; Celtic & 26 \\
        \hline
        \rowcolor{brown!10}\textbf{Total} & \textbf{1013} \\
        \hline
    \end{tabularx}
    \caption{Culture Sample Counts from the British Isles (Viking Section).}
    \label{tab:culture-sample-viking}
\end{table}

\begin{table}[ht]
    \centering
    \small
    \begin{tabularx}{\columnwidth}{Xl}
        \hline
        \rowcolor{brown!30} \textbf{Place} & Central America \\
        \rowcolor{brown!20} \textbf{Section} & Maya \\
        \hline
        \rowcolor{brown!10} \textbf{Culture} & \textbf{Samples} \\
        \hline
        Classic Maya; Classic & 3 \\
        Classic Maya; Late Preclassic Maya & 64 \\
        Formative (Pre-Classic); Early Classic Maya & 8 \\
        Late Classic Maya & 23 \\
        Olmec; Maya & 1 \\
        Classic Maya & 275 \\
        Preclassic Maya & 10 \\
        Classic Maya; Late Classic & 2 \\
        Classic Maya; Olmec & 1 \\
        Preclassic Maya; Classic Maya & 2 \\
        Maya & 95 \\
        Late Classic Maya; Late Classic & 4 \\
        \hline
        \rowcolor{brown!10}\textbf{Total} & \textbf{488} \\
        \hline
    \end{tabularx}
    \caption{Culture Sample Counts from Central America (Maya Section).}
    \label{tab:culture-sample-maya}
\end{table}

\begin{table}[ht]
    \centering
    \small
    \begin{tabularx}{\columnwidth}{Xl}
        \hline
        \rowcolor{brown!30} \textbf{Place} & Egypt \\
        \rowcolor{brown!20} \textbf{Section} & Ancient Egyptian \\
        \hline
        \rowcolor{brown!10} \textbf{Culture} & \textbf{Samples} \\
        \hline
        6$^{th}$ Dynasty & 1 \\
        Late Cypriot; 18$^{th}$ Dynasty & 1 \\
        26$^{th}$ Dynasty; Archaic Greek; Punic & 1 \\
        Late Period; 30$^{th}$ Dynasty & 1 \\
        30$^{th}$ Dynasty; Ptolemaic & 15 \\
        22$^{nd}$ Dynasty & 69 \\
        18$^{th}$ Dynasty; 19$^{th}$ Dynasty & 2 \\
        New Kingdom; 19$^{th}$ Dynasty; 20$^{th}$ Dynasty & 1 \\
        12$^{th}$ Dynasty & 1 \\
        26$^{th}$ Dynasty; Archaic Greek; East Greek; Hellenistic & 1 \\
        New Kingdom & 21 \\
        Late Predynastic; 1$^{st}$ Dynasty & 2 \\
        25$^{th}$ Dynasty & 7 \\
        30$^{th}$ Dynasty & 128 \\
        Middle Kingdom & 1 \\
        Late Period & 96 \\
        18$^{th}$ Dynasty; 21St Dynasty & 1 \\
        21$^{st}$  Dynasty & 171 \\
        19$^{th}$ Dynasty; 20$^{th}$ Dynasty & 3 \\
        20$^{th}$ Dynasty; 21St Dynasty & 2 \\
        26$^{th}$ Dynasty & 257 \\
        19$^{th}$ Dynasty & 40 \\
        18$^{th}$ Dynasty & 95 \\
        1$^{st}$  Dynasty & 7 \\
        Ramesside & 21 \\
        23$^{rd}$  Dynasty & 1 \\
        22$^{nd}$  Dynasty; 23R$^{rd}$ Dynasty & 3 \\
        26$^{th}$ Dynasty; 27$^{th}$ Dynasty & 1 \\
        20$^{th}$ Dynasty & 25 \\
        Late Period; 26$^{th}$ Dynasty & 3 \\
        25$^{th}$ Dynasty; Kushite & 1 \\
        26$^{th}$ Dynasty; Punic; Archaic Greek & 1 \\
        27$^{th}$ Dynasty & 13 \\
        25$^{th}$ Dynasty; 26$^{th}$ Dynasty & 1 \\
        Third Intermediate & 47 \\
        Late Period; Archaic Greek & 2 \\
        Late Period; Ptolemaic & 5 \\
        29$^{th}$ Dynasty & 1 \\
        New Kingdom; Third Intermediate & 1 \\
        Ancient Egypt & 5 \\
        26$^{th}$ Dynasty; 30$^{th}$ Dynasty & 1 \\
        \hline
        \rowcolor{brown!10}\textbf{Total} & \textbf{1056} \\
        \hline
    \end{tabularx}
    \caption{Culture Sample Counts from Egypt (Ancient Egyptian Section).}
    \label{tab:culture-sample-egypt}
\end{table}

\begin{table}[ht]
    \centering
    \small
    \begin{tabularx}{\columnwidth}{Xl}
        \hline
        \rowcolor{brown!30}\textbf{Place} & India \\
        \rowcolor{brown!20}\textbf{Section} & Mohenjo-Daro \\
        \hline
        \rowcolor{brown!10}\textbf{Culture} & \textbf{Sample Count} \\
        \hline
        Indus Valley Civil. & 114 \\
          \hline
          & \\
           \hline
          \rowcolor{brown!20}\textbf{Section} & Mauryan \\
     
         \rowcolor{brown!10}\textbf{Culture} & \textbf{Sample Count} \\
        \hline
        Mauryan & 17 \\
        \hline
         & \\
        \hline
        \rowcolor{brown!20}\textbf{Section} & Gupta Dynasty \\
    
        \rowcolor{brown!10}\textbf{Culture} & \textbf{Sample Count} \\
             \hline
        Gupta & 737 \\
        \hline
          \rowcolor{brown!10}\textbf{Total}&\textbf{868}\\
         \hline
    \end{tabularx}
    \caption{Culture Sample Counts from India.}
    \label{tab:tab:culture-sample-ind}
\end{table}

\begin{table}[ht]
    \centering
    \small
    \begin{tabularx}{\columnwidth}{Xl}
        \hline
        \rowcolor{brown!30} \textbf{Place} & Iran \\
        \rowcolor{brown!20} \textbf{Section} & Persian \\
        \hline
        \rowcolor{brown!10} \textbf{Culture} & \textbf{Samples} \\
        \hline
        Inju Dynasty & 3 \\
        Middle Islamic; Seljuq Dynasty; Persian & 1 \\
        Safavid Dynasty; Mughal Dynasty & 1 \\
        Persian; Islamic & 11 \\
        Persian; Late Islamic & 3 \\
        Samanid Dynasty & 27 \\
        Safavid Dynasty & 395 \\
        Timurid Dynasty; Islamic & 1 \\
        Safavid Dynasty; Post-Medieval & 1 \\
        Mughal Dynasty; Persian & 1 \\
        Ilkhanid Dynasty; Persian & 3 \\
        Turkman Dynasty & 3 \\
        Early Sasanian; Safavid Dynasty & 1 \\
        Islamic; Safavid Dynasty & 1 \\
        Ilkhanid Dynasty & 192 \\
        Middle Islamic; Persian & 6 \\
        Islamic; Qajar Dynasty & 2 \\
        Persian; Safavid Dynasty & 1 \\
        Safavid Dynasty; Persian; Islamic & 2 \\
        Mughal Dynasty; Safavid Dynasty & 1 \\
        Qajar Dynasty & 193 \\
        Safavid Dynasty; Islamic & 4 \\
        Persian; Mughal Dynasty & 1 \\
        Islamic; Persian & 2 \\
        Timurid Dynasty & 35 \\
        Persian & 108 \\
        \hline
        \rowcolor{brown!10}\textbf{Total} & \textbf{999} \\
        \hline
    \end{tabularx}
    \caption{Culture Sample Counts from Iran (Persian Section).}
    \label{tab:culture-sample-persian}
\end{table}

\begin{table}[ht!]
    \centering
    \small
    \begin{tabularx}{\columnwidth}{Xl}
        \hline
        \rowcolor{brown!30} \textbf{Place} & China \\
        \rowcolor{brown!20} \textbf{Section} & Tang Dynasty \\
        \hline
        \rowcolor{brown!10} \textbf{Culture} & \textbf{Samples} \\
        \hline
        Tang Dynasty; Sui Dynasty & 1 \\
        Tang Dynasty; Ming Dynasty & 3 \\
        Tang Dynasty; Ming Dynasty; Jin Dynasty; Yuan Dynasty & 1 \\
        Tang Dynasty; Song Dynasty & 1 \\
        Song Dynasty; Tang Dynasty & 1 \\
        Liao Dynasty; Tang Dynasty & 2 \\
        Tang Dynasty; Northern Wei Dynasty & 1 \\
        Six Dynasties; Tang Dynasty & 5 \\
        Tang Dynasty & 1 \\
        Northern Qi Dynasty; Sui Dynasty; Tang Dynasty & 1 \\
        Tang Dynasty; Liao Dynasty & 3 \\
        Six Dynasties; Sui Dynasty; Tang Dynasty & 1 \\
        Tang Dynasty; Five Dynasties; Northern Song Dynasty & 381 \\
        Five Dynasties; Tang Dynasty & 4 \\
        Tang Dynasty & 628 \\
        Sui Dynasty; Tang Dynasty & 5 \\
        \hline
        \rowcolor{brown!10}\textbf{Total} & \textbf{1039} \\
        \hline
    \end{tabularx}
    \caption{Culture Sample Counts from China (Tang Dynasty Section).}
    \label{tab:culture-sample-tang}
\end{table}

\begin{table}[t!]
    \centering
    \small
    \begin{tabularx}{\columnwidth}{Xl}
        \hline
        \rowcolor{brown!30} \textbf{Place} & Japan \\
        \rowcolor{brown!20} \textbf{Section} & Japanese \\
        \hline
        \rowcolor{brown!10} \textbf{Culture} & \textbf{Samples} \\
        \hline
        Momoyama Period & 6 \\
        Genroku Era; Hoei Era & 1 \\
        Asuka Period & 1 \\
        Muromachi Period; Momoyama Period & 2 \\
        Late Kofun; Nara Period & 1 \\
        Nara Period & 12 \\
        Middle Kofun & 13 \\
        Yayoi Period & 5 \\
        Middle Kofun; Late Kofun & 34 \\
        Edo Period; Kamakura Period & 1 \\
        Oei Era & 2 \\
        Kyowa Era; Oei Era & 1 \\
        Edo Period; Momoyama Period & 1 \\
        Jomon Period & 16 \\
        Kyowa Era & 1 \\
        Bunka Era & 1 \\
        Bun'An Era; Bunsei Era & 1 \\
        Muromachi Period & 40 \\
        Asuka Period; Nara Period & 1 \\
        Heian Period & 9 \\
        Muromachi Period; Momoyama Period; Edo Period & 1 \\
        Muromachi Period; Buddhist & 1 \\
        Meiji Era & 1 \\
        Hakuho Period & 1 \\
        Showa Era & 13 \\
        Early Kofun; Middle Kofun & 26 \\
        Nanbokucho Period & 2 \\
        Kofun Period; Edo Period & 1 \\
        Edo Period & 24 \\
        Kamakura Period; Meiji Era & 1 \\
        Kofun Period & 419 \\
        Early Kofun & 7 \\
        Wado Era & 1 \\
        Late Kofun & 179 \\
        Kofun Period; Asuka Period & 5 \\
        Kamakura Period & 26 \\
        Nara Period; Edo Period & 1 \\
        Kofun Period; Nara Period & 1 \\
        Kamakura Period; Muromachi Period & 9 \\
        Heian Period; Kamakura Period & 1 \\
        \hline
        \rowcolor{brown!10}\textbf{Total} & \textbf{869} \\
        \hline
    \end{tabularx}
    \caption{Culture Sample Counts from Japan (Japanese Section).}
    \label{tab:culture-sample-japan}
\end{table}

\begin{table}[ht!]
    \centering
    \small
    \begin{tabularx}{\columnwidth}{Xl}
        \hline
        \rowcolor{brown!30} \textbf{Place} & Iraq \\
        \rowcolor{brown!20} \textbf{Section} & Mesopotamian \\
        \hline
        \rowcolor{brown!10} \textbf{Culture} & \textbf{Samples} \\
        \hline
        Neo-Assyrian; Late Babylonian & 9 \\
        Late Babylonian; Assyrian & 1 \\
        Elamite; Third Dynasty Of Ur & 1 \\
        Early Dynastic (Middle East) & 1 \\
        Old Assyrian; Early Bronze Age III & 1 \\
        Late Uruk & 26 \\
        Isin-Larsa & 3 \\
        Neo-Assyrian & 406 \\
        Uruk & 3 \\
        Late Uruk; Chalcolithic & 1 \\
        Middle Babylonian; Neo-Babylonian Dynasty & 1 \\
        Old Babylonian; Cypriot & 1 \\
        Late Babylonian & 20 \\
        Babylonian; Neo-Assyrian & 1 \\
        Neo-Assyrian; Babylonian & 1 \\
        Assyrian; Late Babylonian & 2 \\
        Jemdet Nasr; Proto-Elamite & 1 \\
        Halaf & 38 \\
        Assyrian & 7 \\
        Middle Assyrian & 11 \\
        Jemdet Nasr & 27 \\
        Third Dynasty Of Ur; Ubaid & 1 \\
        Old Babylonian & 41 \\
        Kassite & 4 \\
        Babylonian & 3 \\
        Neo-Babylonian Dynasty & 2 \\
        Babylonian; Akkadian & 1 \\
        Old Assyrian & 2 \\
        Old Babylonian; Third Dynasty Of Ur & 1 \\
        Ubaid & 15 \\
        Early Dynastic (Middle East); Akkadian & 2 \\
        Early Dynastic II & 2 \\
        Isin-Larsa; Old Babylonian & 1 \\
        Jemdet Nasr; Akkadian & 1 \\
        Old Babylonian; Assyrian & 1 \\
        Akkadian & 102 \\
        Early Dynastic III; Akkadian & 10 \\
        Old Babylonian; Old Assyrian & 1 \\
        Isin-Larsa; Old Babylonian; Kassite & 1 \\
        Uruk; Jemdet Nasr & 3 \\
        Early Dynastic II; Early Dynastic III & 1 \\
        Early Dynastic III & 81 \\
        Mesopotamian & 1 \\
        Late Babylonian; Neo-Assyrian & 3 \\
        Assyrian; Ubaid & 1 \\
        Third Dynasty Of Ur; Old Babylonian & 2 \\
        Third Dynasty Of Ur & 137 \\
        Neo-Assyrian; Phoenician & 2 \\
        Middle Babylonian & 11 \\
        Lagash II & 1 \\
        Third Dynasty Of Ur; Isin-Larsa & 4 \\
        \hline
        \rowcolor{brown!10}\textbf{Total} & \textbf{1000} \\
        \hline
    \end{tabularx}
    \caption{Culture Sample Counts from Iraq (Mesopotamian Section).}
    \label{tab:culture-sample-iraq}
\end{table}

\section{TimeTravel Benchmark Examples}
\label{appendix:data_stats}

\begin{figure}[!t]
\begin{tcolorbox}[colback=brown!5, colframe=brown!35,
    fonttitle=\bfseries,
    ]
\small

\lstset{breaklines=true, basicstyle=\scriptsize\ttfamily} 
\begin{center}\includegraphics[width=0.75\linewidth]{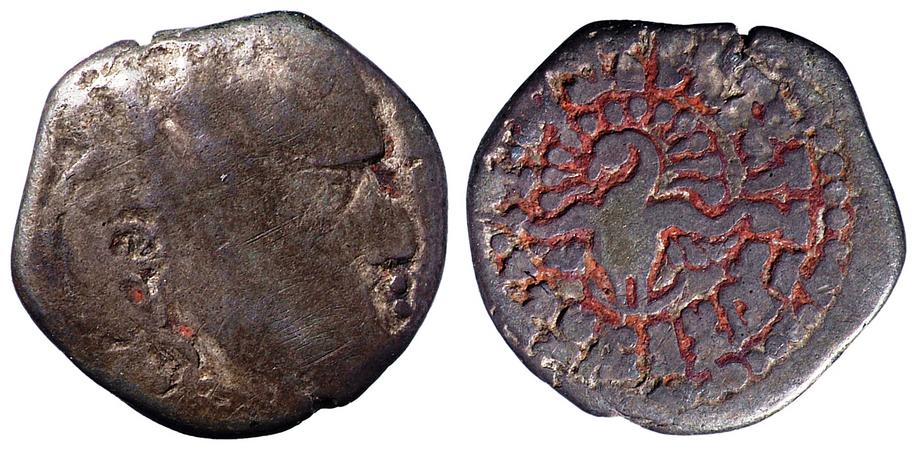}\end{center}
\begin{lstlisting}
{
    "id": 4621,
    "Image": "../Documents/2014_11/5_17/e572b5a1_adcb_4691_871c_a3da0123a3fa/preview_01222265_001.jpg",
    "Description": "Silver coin.; Bust of Skandagupta, facing to the right, wearing a decorated cap and collar.; Peacock, with outspread wings and tail, standing facing with head to left. Surrounded by a border of dots.",
    "Production date": "456-467 (about)",
    "Find spot": null,
    "Materials": "silver",
    Technique": null,
    "Inscription": "Inscription type: inscription Inscription position: reverse Inscription language: Sanskrit Inscription script: Brahmi",
    "Subjects": "king/queen; peacock",
    "Assoc name": "Named in inscription & portrayed: Skandagupta",
    "Culture": "Gupta",
    "Section": "Gupta dynasty",
    "Place": "India"
 }
 
 for this artifact, we generated a description such as:
{
 "description": "This silver coin from the Gupta dynasty, produced approximately between 456-467 CE, features a distinguished portrait of Skandagupta on the obverse. Wearing an ornate cap and collar, Skandagupta is depicted facing right, encapsulating the regal aesthetic of the period. The reverse displays a majestic peacock, wings and tail fully spread, elegantly poised with its head turned to the left, bordered by a circle of dots adding to the coin's decorative appeal. An inscription in Sanskrit, using the Brahmi script, adorns the reverse, providing linguistic and cultural context. This artifact, emblematic of Gupta artistry and power, underscores the dynastic symbolism and reverence for nature characteristic of this era in Indian history."
}
\end{lstlisting}
\end{tcolorbox}
\caption{
\small
This entry represents a silver coin from the Gupta dynastyfrom India, featuring a distinguished portrait of Skandagupta on the obverse. GPT-4o generated a detailed, context-aware description based on the available metadata, highlighting its craftsmanship, ceremonial significance, and cultural context.}
\label{fig:raw-data-india}
\end{figure}

\begin{figure}[!t]
\begin{tcolorbox}[colback=brown!5, colframe=brown!35,
    fonttitle=\bfseries,
    ]
\small

\lstset{breaklines=true, basicstyle=\scriptsize\ttfamily} 
\begin{center}\includegraphics[width=0.75\linewidth]{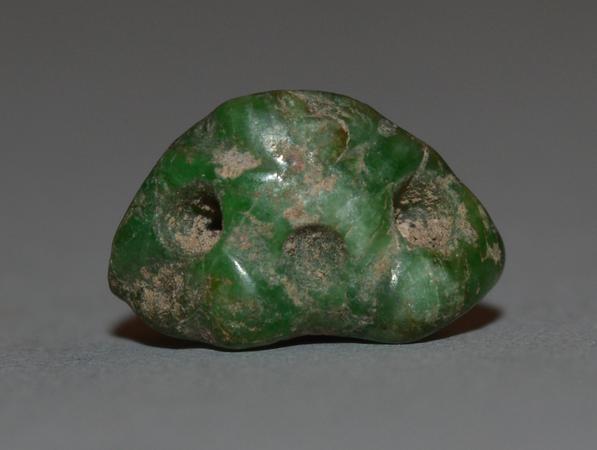}\end{center}
\begin{lstlisting}
  {
  
    "id": 7443,
    "Image": "../Documents/2015_9/4_10/b0f2152e_0ab5_4585_a7ca_a50900a611e8/preview_1997_q_1879__a_.jpg",
    "Description": "Votive object in the form of polished piece of jade with six holes drilled in it.",
    "Production date": null,
    "Find spot": "Excavated/Findspot: Tzimin Kax (?);  Excavated/Findspot: Hatzcap Ceel (?);  Excavated/Findspot: Cahal Pichik (?)",
    "Materials": "jade",
    "Technique": "polished; drilled",
    "Inscription": null,
    "Subjects": null,
    "Assoc name": null,
    "Culture": "Classic Maya; Late Preclassic Maya",
    "Section": "Maya",
    "Place": "Central America"
  }
    for this artifact, we generated a description such as:
{
 "description": "This votive object, crafted by the Classic and Late Preclassic Maya cultures, is a testament to the artistry and spiritual practices of ancient Central America. Made from polished jade, a material revered for its symbolic and aesthetic value, this artifact features six precision-drilled holes, showcasing the sophisticated techniques employed by the Maya artisans. Although specific inscription details are absent, the object's polished surface and drilled holes suggest it may have had ceremonial or religious significance. Unearthed at find spots such as Tzimin Kax, Hatzcap Ceel, and Cahal Pichik, this piece provides valuable insights into the ritualistic and cultural engagements of the Maya civilization."
 
}
\end{lstlisting}
\end{tcolorbox}
\caption{
\small
This particular entry represents a polished jade votive object from the Classic and Late Preclassic Maya features six precision-drilled holes, reflecting advanced craftsmanship and likely ceremonial significance. Unearthed at sites like Tzimin Kax, it offers insight into Maya rituals.}
\label{fig:raw-data-maya}
\end{figure}

\begin{figure*}[t!]
    \centering
    \includegraphics[width=\textwidth]{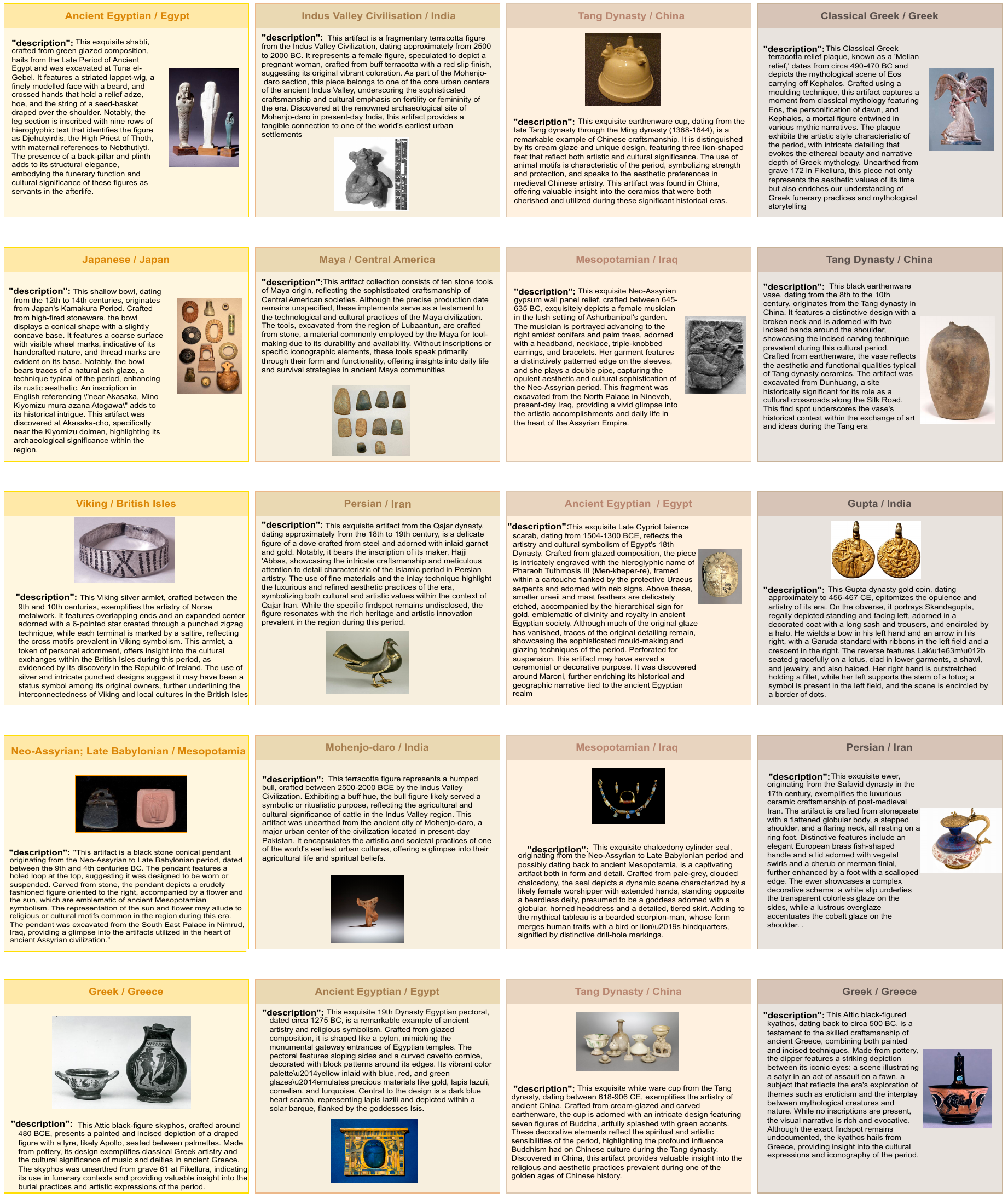}
    \caption{Cultural and material diversity of TimeTravel dataset samples across civilizations and historical periods. The dataset includes artifacts from Ancient Egypt, Greece, Mesopotamia, China, and Japan, spanning prehistoric to medieval times. A wide range of materials, including ceramics, metals, and stone, highlights artistic, technological, and societal influences, ensuring a comprehensive representation of historical craftsmanship and cultural heritage.}
    \label{fig:appendix_qual_examples}
\end{figure*}

\begin{figure*}[t!]
    \centering
    \includegraphics[width=\textwidth]{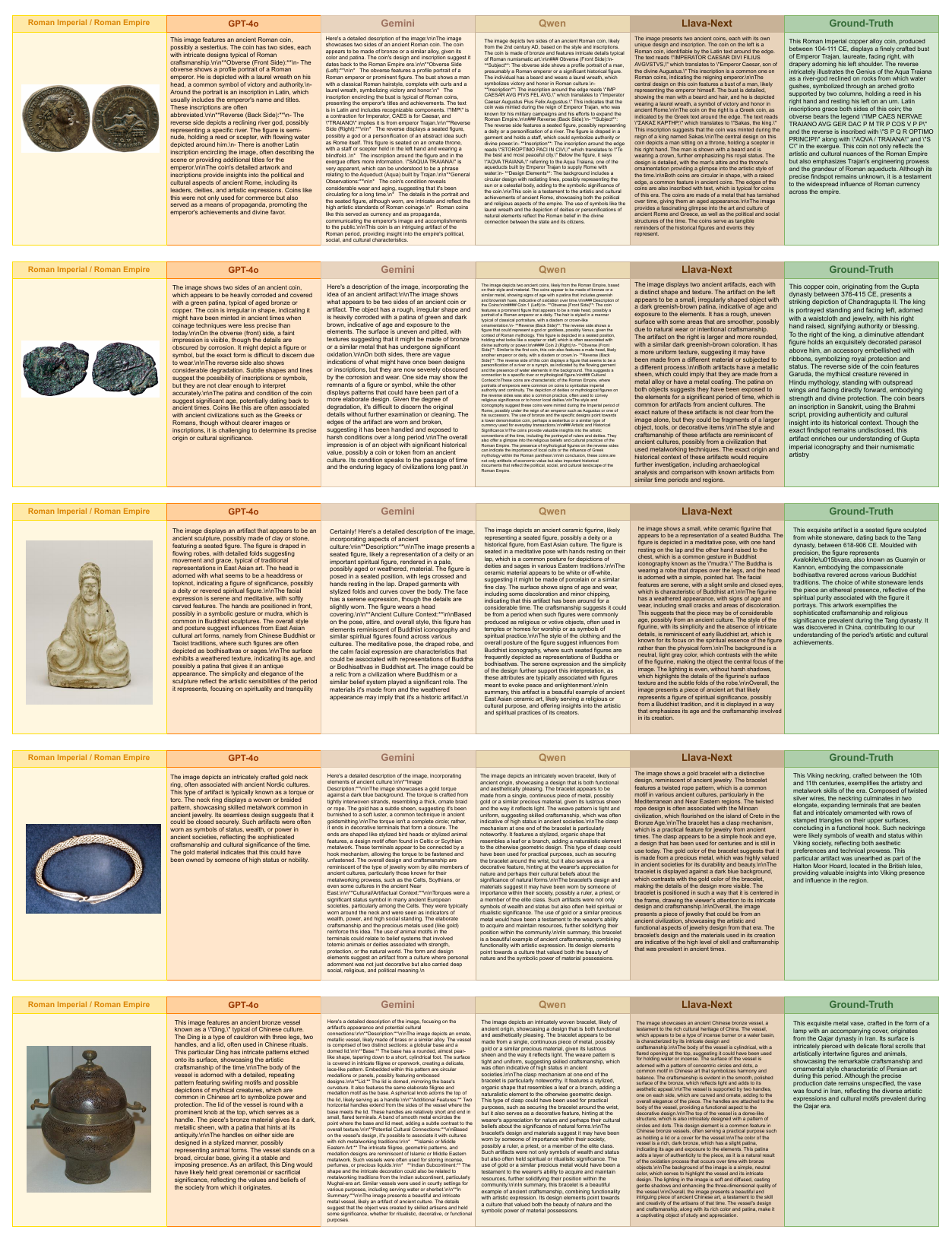}
    \caption{ Cross-model comparison of generated descriptions for TimeTravel dataset samples, highlighting variations in detail and accuracy. It illustrates differences in descriptive depth across open- and closed-source models, emphasizing the diversity in interpretative approaches and alignment with the ground truth.}
    \label{fig:appendix_data_examples}
\end{figure*}

\end{document}